\documentclass[11pt]{article}
\usepackage[preprint]{acl}
\usepackage{times}
\usepackage{latexsym}
\usepackage[T1]{fontenc}
\usepackage[utf8]{inputenc}
\usepackage{microtype}
\usepackage{graphicx}
\usepackage{booktabs}

\title{Specialized Decision Models vs.\ General-Purpose LLMs:\\Benchmarking Jev Across Knowledge, Reasoning, and Multilingual Tasks}

\author{
  \textbf{Xing Li\textsuperscript{1}\thanks{Equal contribution.},
  Qingcheng Chang\textsuperscript{1}\footnotemark[1],
  Jinzhong Ning\textsuperscript{1}\thanks{Corresponding author.},
  Changfeng Xu\textsuperscript{1}}\\
  \textbf{Shenlong Zhang\textsuperscript{1},
  Yijia Zhang\textsuperscript{1},
  Ling Luo\textsuperscript{2},
  Hongfei Lin\textsuperscript{2}}\\
  \textsuperscript{1}Dalian Maritime University\\
  \textsuperscript{2}Dalian University of Technology\\
  \texttt{lllxxx@dlmu.edu.cn, cqc@dlmu.edu.cn, ningjinzhong@dlmu.edu.cn}\\
  \texttt{xuchangfeng@dlmu.edu.cn, zhangshenlong@dlmu.edu.cn}\\
  \texttt{lingluo\_nlp@outlook.com, hflin@dlut.edu.cn}
}

\begin{document}
\maketitle

\begin{abstract}
Jev is a ``System One'' model that returns a choice among given options instead of generating text. We study how such a specialized decision model compares with general-purpose large language models (LLMs). We evaluate Jev on 13 multiple-choice benchmarks covering knowledge, reasoning, and multilingual understanding, and compare it with 19 LLMs in three tiers: frontier, representative, and small. Jev is competitive with frontier LLMs on knowledge and commonsense benchmarks and obtains the best score on MMLU-Redux and ARC-Challenge. Outside mathematics, it also outperforms most representative LLMs and all small LLMs. However, it falls behind on mathematical word problems: on MathQA, it is 17.7 points below the frontier median and scores lower than all 19 LLMs. These results indicate that a specialized decision model can match general-purpose LLMs on decisions that rely mainly on knowledge, but not on decisions that require multi-step calculation.
\end{abstract}

\suppressfloats[t]
\begin{table}[t]
  \centering
  \small
  \begin{tabular}{llcc}
    \toprule
    \textbf{Category} & \textbf{Benchmark} & \textbf{Lang.} & \textbf{Options} \\
    \midrule
    Knowledge & MMLU-Pro & En & 10 \\
     & MMLU-Redux & En & 4 \\
     & MMLU & En & 4 \\
     & GPQA Diamond & En & 4 \\
     & C-Eval & Zh & 4 \\
     & HellaSwag & En & 4 \\
     & WinoGrande & En & 2 \\
     & TruthfulQA & En & varies \\
    \midrule
    Reasoning & ARC-Challenge & En & 4 \\
     & MathQA & En & 5 \\
     & AQuA-RAT & En & 5 \\
    \midrule
    Multilingual & MMLU-ProX & 29 & 10 \\
     & MMMLU & 14 & 4 \\
    \bottomrule
  \end{tabular}
  \caption{Benchmarks used in this study, with their language (or number of languages) and number of answer options.}
  \label{tab:benchmarks}
\end{table}

\begin{figure*}[t]
  \centering
  \includegraphics[width=\textwidth]{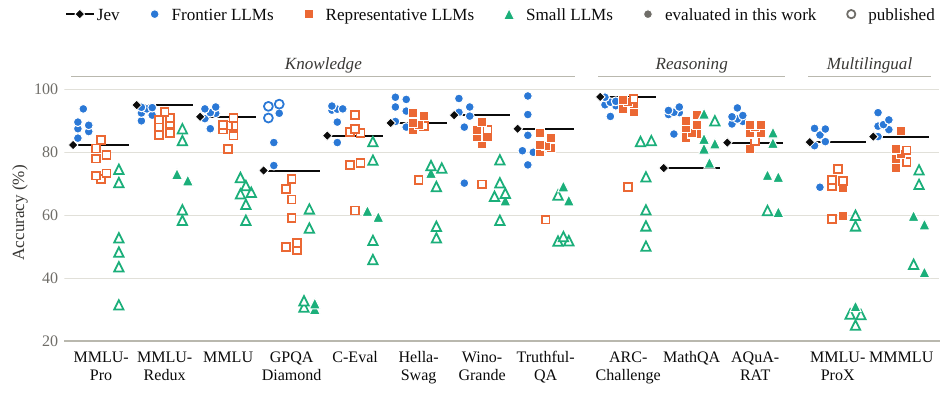}
  \caption{Jev and three tiers of LLMs on 13 benchmarks. Each marker is the score of one model, and the black line marks Jev. Filled markers are scores evaluated in this work; hollow markers are published scores (Appendix~\ref{app:sources}).}
  \label{fig:overview}
\end{figure*}

\section{Introduction}
\label{sec:introduction}

Large language models (LLMs) have achieved strong performance across a wide range of language understanding and reasoning tasks \citep{openai2026gpt56,deepseek2024v3,yang2025qwen3}. Beyond generating text, they are increasingly used to make decisions in applications such as question answering, text classification, information verification, and AI agents. Many of these decisions involve choosing an answer or action from a predefined set of options. Although general-purpose LLMs can handle such tasks effectively, their broad capabilities are not always required by the application. This motivates an important question: to what extent can specialized decision models match general-purpose LLMs on tasks with finite answer spaces, and where do their capabilities begin to diverge?

Jev, developed by TypeSafe AI, provides an opportunity to investigate this question. It is a specialized ``System One'' model designed to make structured decisions rather than generate open-ended text \citep{typesafe2026jev}. Given an input and a set of candidate options, Jev directly selects an answer and returns a probability distribution over the options. Since its release, several studies have explored its performance in text classification, routing, medical and legal applications, model evaluation, and scientific decision making \citep{deusser2026jev,madrid2026medicine,zhang2026same,li2026judge,wang2026trace,deng2026science}. These studies provide initial evidence of Jev's effectiveness in specific settings. However, its broader capability profile remains insufficiently understood. In particular, previous evaluations either focus on individual applications or include a limited range of comparison models. Even the most comprehensive existing study, which evaluates 37 datasets, compares Jev with only two open-weight LLMs \citep{deusser2026jev}. As a result, it remains unclear how Jev compares with LLMs at different capability levels, and whether its relative performance changes across tasks requiring different types of knowledge and reasoning.

In this work, we conduct a systematic empirical evaluation of Jev against general-purpose LLMs on standard multiple-choice benchmarks. We evaluate 13 benchmarks covering general knowledge, commonsense reasoning, scientific question answering, mathematical reasoning, and multilingual understanding (Table~\ref{tab:benchmarks}). To provide a broad comparison, we include 19 LLMs organized into three tiers: current frontier models, representative models from earlier generations, and small models with fewer than 10B parameters (Figure~\ref{fig:overview}). All models are evaluated within a finite-answer setting, allowing us to examine their ability to select correct answers under the same candidate constraints. Rather than focusing only on overall accuracy, we investigate how Jev's relative performance varies across benchmarks with different task demands.

Our results show that Jev exhibits a distinctive capability profile. It performs competitively with general-purpose LLMs on several knowledge-intensive and commonsense benchmarks, achieving the highest accuracy among the evaluated models on MMLU-Redux and ARC-Challenge. It also approaches frontier-level performance on MMLU, WinoGrande, and TruthfulQA. However, this competitiveness does not extend uniformly to all tasks. Jev performs substantially worse on mathematical reasoning benchmarks, falling 17.7 percentage points below the frontier-model median on MathQA and ranking behind all 19 comparison models. These findings suggest that specialized decision models can be effective alternatives to general-purpose LLMs for some constrained decision tasks, but their effectiveness depends strongly on the capabilities required by the task. In particular, selecting from a fixed set of answers does not necessarily make a task easy: some decisions still require substantial calculation and multi-step reasoning.

Our contributions are summarized as follows:
\begin{itemize}
    \item We investigate the capability boundary between specialized decision models and general-purpose LLMs in finite-answer tasks, using Jev as a representative case study.
    \item We systematically compare Jev with 19 LLMs across three capability tiers on 13 standard benchmarks, revealing its strengths and limitations across knowledge, commonsense, mathematical reasoning, and multilingual understanding.
    \item We identify task-dependent performance differences that provide empirical evidence for selecting between specialized decision models and general-purpose LLMs in future decision-making systems.
\end{itemize}

\section{Related Work}

\paragraph{LLM benchmarks.} Multiple-choice benchmarks are the standard tools for measuring the knowledge and commonsense of LLMs. They include MMLU, ARC, and HellaSwag \citep{hendrycks2021mmlu,clark2018arc,zellers2019hellaswag} and harder or cleaner successors such as MMLU-Pro, MMLU-Redux, and GPQA \citep{wang2024mmlupro,gema2025mmluredux,rein2023gpqa}. C-Eval and the multilingual extensions of MMLU cover other languages \citep{huang2023ceval,xuan2025mmluprox,openai2024mmmlu}. We use these benchmarks unchanged and apply them to a new type of model, a specialized decision model.

\paragraph{Jev and its evaluations.} Jev answers typed questions about a given input. Choice selects one of up to 255 options, Score places the input on an ordered rubric, and Noul returns the probability of a yes/no statement \citep{typesafe2026jev,deusser2026jev}. Its documentation lists arithmetic, counting, and multi-hop instructions as known weaknesses \citep{typesafe2026jagged}. Jev has been evaluated on classification and routing \citep{deusser2026jev}, medicine \citep{madrid2026medicine}, legal documents \citep{zhang2026same}, judging \citep{li2026judge,wang2026trace}, scientific decisions \citep{deng2026science}, and social-science annotation \citep{ibrahim2026annotation}. Other work studies its sensitivity to option names \citep{sun2026typesafe} and its application ecosystem \citep{ling2026wild}. These studies address specific applications or compare Jev with only a few models, whereas we compare it with three tiers of general-purpose LLMs on standard benchmarks.

\paragraph{General-purpose and specialized models.} General-purpose LLMs are increasingly used for text classification and annotation \citep{gilardi2023chatgpt}. However, on text classification, smaller models fine-tuned for the task often still outperform zero-shot LLMs \citep{bucher2024finetuned}. Model routing and cascades exploit such differences by sending each query to the cheapest model likely to answer it well \citep{chen2023frugalgpt,ding2024hybrid,ong2024routellm}. These methods choose among general-purpose LLMs of different sizes. We study a different kind of component, a specialized decision model, and examine which constrained decisions it can take over.

\section{Experiments}

\begin{table*}[t]
  \centering
  \footnotesize
  \setlength{\tabcolsep}{3pt}
  \resizebox{\textwidth}{!}{%
  \begin{tabular}{lccccccccccccc}
    \toprule
     & \multicolumn{8}{c}{\textbf{Knowledge}} & \multicolumn{3}{c}{\textbf{Reasoning}} & \multicolumn{2}{c}{\textbf{Multilingual}} \\
    \cmidrule(lr){2-9} \cmidrule(lr){10-12} \cmidrule(lr){13-14}
    \textbf{Model} & \textbf{Pro} & \textbf{Redux} & \textbf{MMLU} & \textbf{GPQA} & \textbf{C-Eval} & \textbf{HSwag} & \textbf{Wino} & \textbf{TQA} & \textbf{ARC-C} & \textbf{MathQA} & \textbf{AQuA} & \textbf{ProX} & \textbf{MMMLU} \\
    \midrule
    \textbf{Jev} & 82.3 & \textbf{95.0} & 91.3 & 74.2 & 85.3 & 89.3 & 91.8 & 87.5 & \textbf{97.6} & 75.0 & 83.1 & 83.2 & 85.0 \\
    \midrule
    \multicolumn{14}{l}{\textit{Frontier LLMs}} \\
    \textbf{GPT-5.6 Sol}$^\dagger$ & 84.5 & 90.0 & 90.7 & \textit{94.6} & 89.6 & 89.8 & 70.2 & 76.0 & 95.1 & 92.0 & 89.0 & 68.9 & 85.0 \\
    \textbf{Claude Opus 5.5}$^\dagger$ & 86.6 & 94.2 & 92.6 & 75.8 & 83.1 & 96.8 & 92.7 & 92.0 & 94.8 & 85.8 & 89.8 & 82.1 & 87.2 \\
    \textbf{Gemini 3.8 Flash}$^\dagger$ & \textbf{93.8} & 93.8 & \textbf{94.4} & \textbf{\textit{95.3}} & \textbf{94.7} & \textbf{97.5} & \textbf{97.1} & \textbf{97.9} & 97.5 & 92.7 & 91.3 & \textbf{87.6} & \textbf{92.6} \\
    \textbf{Qwen3.8-Flash} & 88.6 & 92.5 & 92.4 & 83.1 & 93.7 & 94.4 & 94.4 & 85.4 & 96.2 & 93.3 & 91.7 & 83.4 & 88.3 \\
    \textbf{Qwen3.8-Max} & 89.6 & 94.3 & 93.8 & 92.4 & 93.4 & 88.0 & 88.0 & 80.5 & 91.4 & \textbf{94.4} & \textbf{94.1} & 87.4 & 90.3 \\
    \textbf{DeepSeek-V4.1-Flash} & 87.5 & 91.8 & 87.5 & \textit{90.9} & 93.8 & 93.1 & 91.5 & 80.0 & 95.9 & 92.6 & 90.6 & 85.5 & 88.8 \\
    \midrule
    \multicolumn{14}{l}{\textit{Representative LLMs}} \\
    \textbf{GPT-4o}$^\dagger$ & \textit{72.6} & \textit{88.0} & \textit{87.2} & \textit{49.9} & \textit{76.0} & 92.6 & 87.1 & 86.0 & 96.1 & 86.0 & 88.0 & \textit{71.1} & 77.9 \\
    \textbf{Claude 3.5 Sonnet}$^\dagger$ & \textit{78.0} & \textit{88.9} & \textit{88.3} & \textit{65.0} & \textit{76.7} & 88.6 & 85.0 & 81.5 & 93.7 & 90.0 & 86.0 & 68.7 & 81.1 \\
    \textbf{DeepSeek-R1} & \textit{84.0} & \textit{92.9} & \textit{90.8} & \textit{71.5} & \textit{91.8} & 91.5 & 89.7 & 84.5 & 96.6 & 88.8 & 88.8 & \textit{74.6} & 86.8 \\
    \textbf{DeepSeek-V3} & \textit{81.2} & \textit{90.4} & \textit{88.5} & \textit{59.1} & \textit{86.5} & 89.2 & 84.9 & 82.1 & 95.3 & 87.8 & 86.3 & \textit{70.8} & 79.4 \\
    \textbf{Qwen3-32B} & \textit{79.1} & \textit{90.9} & \textit{81.0} & \textit{68.4} & \textit{87.3} & \textit{71.1} & \textit{69.8} & \textit{58.6}$^{\ast}$ & \textit{69.0} & 91.9 & 88.6 & \textit{69.2} & \textit{80.6} \\
    \textbf{Qwen2.5-72B-Inst.} & \textit{71.6} & \textit{85.6} & \textit{85.3} & \textit{49.0} & \textit{86.1} & 87.2 & 82.7 & 80.1 & 92.7 & 85.7 & \textit{83.7} & \textit{58.7} & \textit{76.9} \\
    \textbf{Llama-3.1-405B-Inst.} & \textit{73.3} & \textit{86.2} & \textit{88.6} & \textit{51.1} & \textit{61.5} & \textit{88.3} & \textit{87.2} & 82.4 & \textit{96.9} & 84.7 & 81.0 & 59.8 & 75.0 \\
    \midrule
    \multicolumn{14}{l}{\textit{Small LLMs}} \\
    \textbf{Qwen3-8B} & \textit{74.6} & \textit{87.5} & \textit{72.0} & \textit{62.0} & \textit{83.4} & \textit{56.5} & \textit{66.0} & \textit{53.2}$^{\ast}$ & \textit{61.7} & 92.3 & 86.3 & \textit{60.0} & \textit{74.4} \\
    \textbf{Qwen3-4B} & \textit{70.4} & \textit{83.7} & \textit{66.8} & \textit{55.9} & \textit{77.5} & \textit{52.8} & \textit{58.4} & \textit{51.8}$^{\ast}$ & \textit{50.2} & \textit{90.0} & 83.0 & \textit{56.6} & \textit{69.8} \\
    \textbf{Llama-3.1-8B-Inst.} & \textit{48.3} & \textit{61.7} & \textit{69.4} & \textit{32.8} & \textit{52.0} & 73.5 & 64.7 & 69.2 & \textit{83.4} & 81.2 & \textit{61.5} & \textit{28.6} & \textit{44.4} \\
    \textbf{Phi-4-mini-inst.} & \textit{52.8} & 73.1 & \textit{67.3} & 32.0 & 61.4 & \textit{69.1} & \textit{67.0} & \textit{66.4}$^{\ast}$ & \textit{83.7} & 82.8 & 72.8 & \textit{28.4} & 59.8 \\
    \textbf{Gemma-3-4B-IT} & \textit{43.6} & 71.1 & \textit{58.4} & \textit{30.8} & 59.5 & \textit{75.0} & \textit{70.3} & \textit{51.9}$^{\ast}$ & \textit{56.6} & 84.3 & 72.2 & \textit{25.1} & 57.1 \\
    \textbf{Mistral-7B-v0.3} & \textit{31.5} & \textit{58.4} & \textit{63.5} & 30.2 & \textit{45.9} & \textit{75.8} & \textit{77.6} & 64.7 & \textit{72.2} & 76.7 & 61.0 & 31.1 & 41.9 \\
    \bottomrule
  \end{tabular}}
  \caption{Accuracy (\%) of Jev and 19 LLMs across 13 benchmarks, grouped into three model tiers. Scores in \textit{italics} are published scores; all others are evaluated in this work. \textbf{Bold} marks the best score in each column. $^{\ast}$Reported with the MC2 metric. $^\dagger$Closed-source. Pro = MMLU-Pro; Redux = MMLU-Redux; GPQA = GPQA Diamond; HSwag = HellaSwag; Wino = WinoGrande; TQA = TruthfulQA; ARC-C = ARC-Challenge; AQuA = AQuA-RAT; ProX = MMLU-ProX.}
  \label{tab:main}
\end{table*}

\subsection{Experimental Setup}

\paragraph{Benchmarks.} We use 13 multiple-choice benchmarks in three categories (Table~\ref{tab:benchmarks}). The knowledge category contains eight benchmarks. \textbf{MMLU} \citep{hendrycks2021mmlu} tests academic knowledge in 57 subjects. \textbf{MMLU-Redux} \citep{gema2025mmluredux} is a re-annotated subset of MMLU with corrected labels. \textbf{MMLU-Pro} \citep{wang2024mmlupro} makes MMLU harder with ten options and more reasoning-focused questions. \textbf{GPQA Diamond} \citep{rein2023gpqa} contains graduate-level science questions written by domain experts. \textbf{C-Eval} \citep{huang2023ceval} covers 52 disciplines in Chinese. \textbf{HellaSwag} \citep{zellers2019hellaswag} asks for the most plausible continuation of a context. \textbf{WinoGrande} \citep{sakaguchi2020winogrande} requires commonsense to resolve an ambiguous reference. \textbf{TruthfulQA} \citep{lin2022truthfulqa} contains questions that people often answer wrongly because of common misconceptions. The reasoning category contains three benchmarks. \textbf{ARC-Challenge} \citep{clark2018arc} consists of grade-school science questions that retrieval-based methods fail to answer. \textbf{MathQA} \citep{amini2019mathqa} and \textbf{AQuA-RAT} \citep{ling2017aqua} consist of math word problems that require multi-step calculation. The multilingual category contains \textbf{MMLU-ProX} \citep{xuan2025mmluprox} and \textbf{MMMLU} \citep{openai2024mmmlu}, which translate MMLU-Pro and MMLU into 29 and 14 languages, respectively.

\paragraph{Models.} We compare Jev with 19 LLMs in three tiers. \textbf{Frontier LLMs} are recently released models: GPT-5.6 Sol, Claude Opus 5.5, Gemini 3.8 Flash, Qwen3.8-Flash, Qwen3.8-Max, and DeepSeek-V4.1-Flash. \textbf{Representative LLMs} are widely used earlier models: GPT-4o, Claude 3.5 Sonnet, DeepSeek-R1, DeepSeek-V3, Qwen3-32B, Qwen2.5-72B-Instruct, and Llama-3.1-405B-Instruct. \textbf{Small LLMs} have fewer than 10B parameters: Qwen3-8B, Qwen3-4B, Llama-3.1-8B-Instruct, Phi-4-mini-instruct, Gemma-3-4B-IT, and Mistral-7B-v0.3.

\paragraph{Evaluation protocol.} We follow the standard evaluation split of each benchmark and report accuracy (\%). For each item, Jev receives the question and the candidate answers as a single Choice request, and its selected option is taken as the prediction. We evaluate Jev and the frontier LLMs ourselves, each with a single run. On GPQA Diamond, the scores of three frontier LLMs are taken from vendor reports and marked as such. The 13 representative and small LLMs and the 13 benchmarks form 169 model--benchmark pairs. For 104 of them, we use scores published by model developers or independent evaluators, and we evaluate the other 65 ourselves. Published scores differ in prompts, few-shot examples, and reasoning settings, so we use them only for rough comparison. Appendix~\ref{app:sources} gives the source of each one. To compare Jev with the frontier tier on a single benchmark, we use the median score of the six frontier LLMs. Appendix~\ref{app:details} gives further details.

\subsection{Results}

\begin{figure}[t]
  \centering
  \includegraphics[width=\columnwidth]{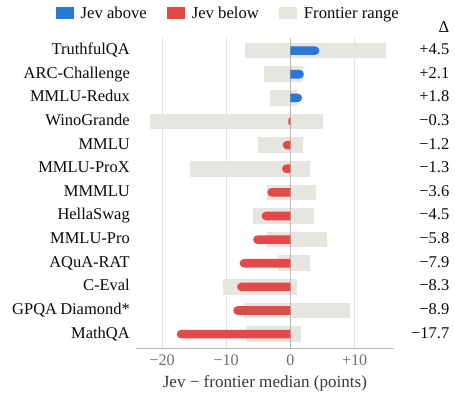}
  \caption{Difference between Jev and the frontier median on each benchmark. Gray bands show the range of the six frontier LLMs. $^{\ast}$On GPQA Diamond, the median and range cover only the three frontier LLMs evaluated in this work.}
  \label{fig:gap}
\end{figure}

Table~\ref{tab:main} reports the accuracy of Jev and the 19 LLMs, and Figure~\ref{fig:gap} shows the difference between Jev and the frontier median on each benchmark. We discuss five contrasts that stand out in these results.

\paragraph{Knowledge versus math.} Jev is competitive with frontier LLMs on knowledge and commonsense benchmarks but falls behind them on math word problems. It obtains the best score among all models on MMLU-Redux and ARC-Challenge. On TruthfulQA, WinoGrande, and MMLU, it is above the frontier median or within about one point of it. However, it scores below all six frontier LLMs on both MathQA and AQuA-RAT, and it trails their median by 17.7 points on MathQA. This contrast reflects the different demands of the two kinds of tasks. A knowledge or commonsense question can be answered by identifying the correct option among the candidates. A math word problem requires several calculation steps before an option can be chosen. A plausible explanation is that general-purpose LLMs can work through these steps in generated text, whereas Jev selects an option directly. Consistent with this, the developer of Jev lists arithmetic among its known weaknesses \citep{typesafe2026jagged}.

\paragraph{Knowledge coverage.} Outside math, Jev is ahead of the representative and small LLMs in almost every comparison, yet all 13 of them outperform it on MathQA. It is ahead of the representative LLMs in 68 of 76 such comparisons, with the largest margins on GPQA Diamond and MMLU-ProX. Against the small LLMs, it is ahead in all 62 such comparisons, mostly by a wide margin. On AQuA-RAT, six representative LLMs and Qwen3-8B also outperform Jev. Even Mistral-7B-v0.3, which is at least 13 points behind Jev on every other benchmark, is slightly ahead on MathQA. Most of our benchmarks test knowledge, so the lead outside math suggests that Jev covers more knowledge than the earlier generation of LLMs. Small models have limited capacity to store knowledge, which explains their large gap to Jev. The lead of both groups on math may come from working through the calculation before answering. This again suggests that the weakness of Jev lies in multi-step calculation rather than in knowledge. Even a small general-purpose model can solve math word problems that Jev cannot.

\paragraph{Reasoning demands.} The gap between Jev and the frontier median grows as a benchmark demands more calculation and multi-step reasoning. It is small on MMLU, moderate on MMLU-Pro, and large on GPQA Diamond, AQuA-RAT, and MathQA. On the graduate-level questions of GPQA Diamond, Jev scores about the same as Claude Opus 5.5 but well below the other frontier LLMs. The MMLU family shows this trend most clearly, because MMLU-Pro covers subjects similar to those of MMLU but contains more reasoning-focused questions. \citet{wang2024mmlupro} report that chain-of-thought prompting \citep{wei2022cot} improves GPT-4o by 1.5 points on MMLU but by 19.1 points on MMLU-Pro, so intermediate reasoning matters far more on MMLU-Pro. Consistent with this difference, Jev is 1.2 points below the frontier median on MMLU but 5.8 points below it on MMLU-Pro. The same pattern appears when Jev is used as a judge \citep{li2026judge} and in medicine \citep{madrid2026medicine}. Among the representative LLMs, the strongest is DeepSeek-R1, which is slightly ahead of Jev on average (Table~\ref{tab:rep}). Besides the two math benchmarks, it leads Jev on MMLU-Pro, C-Eval, HellaSwag, and MMMLU. As a reasoning model, it works through every question before answering, which may explain its lead on the reasoning-focused MMLU-Pro. The gaps on C-Eval and HellaSwag have other causes, which we discuss below.

\paragraph{Translated versus native questions.} Jev transfers well to translated questions but falls behind on questions from Chinese exams. From MMLU to its multilingual version MMMLU, the accuracy of Jev drops by about six points, a decrease similar to those of GPT-5.6 Sol and Claude Opus 5.5. On MMLU-ProX, it is about one point below the frontier median and above all representative and small LLMs. Both benchmarks consist of translated questions, so they test whether the same knowledge can be used in other languages. For Jev, it largely can. C-Eval shows a different result: Jev is about eight points below the frontier median and behind all representative LLMs from Chinese developers. C-Eval is built from Chinese exams rather than translated questions, so it requires knowledge specific to the Chinese curriculum. Models from Chinese developers likely cover this knowledge better.

\begin{table}[t]
  \centering
  \small
  \setlength{\tabcolsep}{4pt}
  \begin{tabular}{lccc}
    \toprule
    \textbf{Benchmark} & \textbf{Ours} & \textbf{\citeauthor{deusser2026jev}} & \textbf{Items} \\
    \midrule
    \textbf{MMLU} & 91.3 & 91.8 & 14,042 \\
    \textbf{C-Eval} & 85.3 & 83.9 & 12,342 \\
    \textbf{HellaSwag} & 89.3 & 95.5 & 10,042 \\
    \textbf{WinoGrande} & 91.8 & 91.4 & 1,267 \\
    \textbf{ARC} & 97.6 & 98.8 & 3,548 \\
    \bottomrule
  \end{tabular}
  \caption{Accuracy (\%) of Jev in this work and in \citet{deusser2026jev}, with the number of items in their evaluation. Our ARC score is on ARC-Challenge; theirs is on ARC-Easy and ARC-Challenge.}
  \label{tab:independent}
\end{table}

\paragraph{Agreement with an independent evaluation.} Our scores for Jev agree closely with those of an independent evaluation, except on HellaSwag. Five of our benchmarks are also used by \citet{deusser2026jev}, which lets us compare the two sets of scores (Table~\ref{tab:independent}). The two evaluations agree within 0.5 points on MMLU and WinoGrande and within 1.4 points on C-Eval. Their ARC score, which also covers the easier ARC-Easy set, is slightly higher than our score on ARC-Challenge. On HellaSwag, our score is 6.2 points lower. The most likely cause is that the two evaluations use different request templates: Jev is known to be sensitive to how the options are presented \citep{typesafe2026jagged,sun2026typesafe}. The gap on HellaSwag in Figure~\ref{fig:gap} may therefore partly reflect our template.

\section{Conclusion}

We evaluated Jev, a specialized decision model, against 19 general-purpose LLMs in three tiers on 13 multiple-choice benchmarks. Jev is competitive with frontier LLMs on knowledge and commonsense benchmarks and outperforms most representative LLMs and all small LLMs there. However, it falls behind on mathematical word problems and scores below all 19 LLMs on MathQA. These results suggest that a specialized decision model can handle many constrained decisions on its own, while decisions that require multi-step calculation still call for a general-purpose LLM. In future work, we will evaluate all tiers under one protocol. We will also study whether the option probabilities of Jev can indicate when a question should be passed to a reasoning model.

\bibliography{custom}

@misc{typesafe2026jev,
  author = {{TypeSafe AI}},
  title = {Introducing System One Models \& {Jev}},
  year = {2026},
  howpublished = {Blog post, September 15},
  url = {https://typesafe.ai/blog/introducing-system-one-models-and-jev}
}

@misc{typesafe2026jagged,
  author = {{TypeSafe AI}},
  title = {Jev 1.13 jaggedness},
  year = {2026},
  howpublished = {Documentation, last reviewed October 2},
  url = {https://docs.typesafe.ai/model-jaggedness/jev-1.13}
}

@article{deusser2026jev,
  author = {Deu{\ss}er, Tobias and Sparrenberg, Lorenz and Sifa, Rafet},
  title = {Evaluating and Benchmarking the System One Model {Jev}},
  journal = {arXiv preprint arXiv:2609.37647},
  year = {2026}
}

@article{madrid2026medicine,
  author = {Madrid-Garc{\'i}a, Alfredo and Merino-Barbancho, Beatriz},
  title = {Jev in Medicine: A Benchmark Evaluation. Preliminary Results},
  journal = {arXiv preprint arXiv:2609.34024},
  year = {2026}
}

@article{zhang2026same,
  author = {Zhang, Fan and Chen, Yankai and Xie, Zhuohan and others},
  title = {Same Scores, Different Decisions: Evaluating {JEV} and Language Models for Legal Document Understanding},
  journal = {arXiv preprint arXiv:2609.27678},
  year = {2026}
}

@article{li2026judge,
  author = {Li, Yubo and Miao, Yidi and Krishnan, Ramayya and Padman, Rema},
  title = {{JEV}-as-a-Judge: Accept When Confident, Escalate When Unsure},
  journal = {arXiv preprint arXiv:2609.26550},
  year = {2026}
}

@article{wang2026trace,
  author = {Wang, Zhiqiang and Gao, Yichao},
  title = {{JEV} as a Judge for Agent Trace Security: An Empirical Comparison with Generative {LLM} Judges},
  journal = {arXiv preprint arXiv:2609.34862},
  year = {2026}
}

@article{deng2026science,
  author = {Deng, Boyuan and Fan, Shuyi and Zhang, Hongyang and Xie, Xinhong},
  title = {Jev for Scientific Decisions: Evaluating Semantic Choices and Their Consequences},
  journal = {arXiv preprint arXiv:2609.24965},
  year = {2026}
}

@article{sun2026typesafe,
  author = {Sun and Xu},
  title = {Type-Safe Is Not Error-Free: A Constrained Decision Head Follows the Option Name, Not the Rubric Bound to It},
  journal = {arXiv preprint arXiv:2609.26758},
  year = {2026}
}

@article{ibrahim2026annotation,
  author = {Ibrahim and Zaki},
  title = {Evaluating Decision Models for Text Annotation in Computational Social Science},
  journal = {arXiv preprint arXiv:2609.24574},
  year = {2026}
}

@article{ling2026wild,
  author = {Ling, Guoming and Xue, Muen and Ye, Zijian},
  title = {Jev in the Wild: A Data-Driven Analysis of the {Jev} Model's Functionality, Applications and Ecosystem},
  journal = {arXiv preprint arXiv:2609.30216},
  year = {2026}
}

@inproceedings{wei2022cot,
  author = {Wei, Jason and Wang, Xuezhi and Schuurmans, Dale and Bosma, Maarten and Ichter, Brian and Xia, Fei and Chi, Ed H. and Le, Quoc V. and Zhou, Denny},
  title = {Chain-of-Thought Prompting Elicits Reasoning in Large Language Models},
  booktitle = {Advances in Neural Information Processing Systems},
  year = {2022}
}

@inproceedings{hendrycks2021mmlu,
  author = {Hendrycks, Dan and Burns, Collin and Basart, Steven and Zou, Andy and Mazeika, Mantas and Song, Dawn and Steinhardt, Jacob},
  title = {Measuring Massive Multitask Language Understanding},
  booktitle = {International Conference on Learning Representations},
  year = {2021}
}

@inproceedings{wang2024mmlupro,
  author = {Wang, Yubo and Ma, Xueguang and Zhang, Ge and others},
  title = {{MMLU-Pro}: A More Robust and Challenging Multi-Task Language Understanding Benchmark},
  booktitle = {Advances in Neural Information Processing Systems (Datasets and Benchmarks Track)},
  year = {2024}
}

@inproceedings{gema2025mmluredux,
  author = {Gema, Aryo Pradipta and Leang, Joshua Ong Jun and Hong, Giwon and others},
  title = {Are We Done with {MMLU}?},
  booktitle = {Proceedings of NAACL},
  year = {2025}
}

@article{rein2023gpqa,
  author = {Rein, David and Hou, Betty Li and Stickland, Asa Cooper and others},
  title = {{GPQA}: A Graduate-Level Google-Proof {Q\&A} Benchmark},
  journal = {arXiv preprint arXiv:2311.12022},
  year = {2023}
}

@inproceedings{huang2023ceval,
  author = {Huang, Yuzhen and Bai, Yuzhuo and Zhu, Zhihao and others},
  title = {{C-Eval}: A Multi-Level Multi-Discipline {Chinese} Evaluation Suite for Foundation Models},
  booktitle = {Advances in Neural Information Processing Systems},
  year = {2023}
}

@inproceedings{zellers2019hellaswag,
  author = {Zellers, Rowan and Holtzman, Ari and Bisk, Yonatan and Farhadi, Ali and Choi, Yejin},
  title = {{HellaSwag}: Can a Machine Really Finish Your Sentence?},
  booktitle = {Proceedings of ACL},
  year = {2019}
}

@inproceedings{sakaguchi2020winogrande,
  author = {Sakaguchi, Keisuke and Le Bras, Ronan and Bhagavatula, Chandra and Choi, Yejin},
  title = {{WinoGrande}: An Adversarial Winograd Schema Challenge at Scale},
  booktitle = {Proceedings of AAAI},
  year = {2020}
}

@inproceedings{lin2022truthfulqa,
  author = {Lin, Stephanie and Hilton, Jacob and Evans, Owain},
  title = {{TruthfulQA}: Measuring How Models Mimic Human Falsehoods},
  booktitle = {Proceedings of ACL},
  year = {2022}
}

@article{clark2018arc,
  author = {Clark, Peter and Cowhey, Isaac and Etzioni, Oren and others},
  title = {Think you have Solved Question Answering? {Try ARC}, the {AI2} Reasoning Challenge},
  journal = {arXiv preprint arXiv:1803.05457},
  year = {2018}
}

@inproceedings{amini2019mathqa,
  author = {Amini, Aida and Gabriel, Saadia and Lin, Shanchuan and Koncel-Kedziorski, Rik and Choi, Yejin and Hajishirzi, Hannaneh},
  title = {{MathQA}: Towards Interpretable Math Word Problem Solving with Operation-Based Formalisms},
  booktitle = {Proceedings of NAACL},
  year = {2019}
}

@inproceedings{ling2017aqua,
  author = {Ling, Wang and Yogatama, Dani and Dyer, Chris and Blunsom, Phil},
  title = {Program Induction by Rationale Generation: Learning to Solve and Explain Algebraic Word Problems},
  booktitle = {Proceedings of ACL},
  year = {2017}
}

@inproceedings{xuan2025mmluprox,
  author = {Xuan, Weihao and Yang, Rui and Qi, Heli and others},
  title = {{MMLU-ProX}: A Multilingual Benchmark for Advanced Large Language Model Evaluation},
  booktitle = {Proceedings of EMNLP},
  year = {2025}
}

@misc{openai2024mmmlu,
  author = {{OpenAI}},
  title = {{MMMLU}: Multilingual Massive Multitask Language Understanding},
  year = {2024},
  howpublished = {Dataset},
  url = {https://huggingface.co/datasets/openai/MMMLU}
}

@article{deepseek2024v3,
  author = {{DeepSeek-AI}},
  title = {{DeepSeek-V3} Technical Report},
  journal = {arXiv preprint arXiv:2412.19437},
  year = {2024}
}

@article{deepseek2025r1,
  author = {{DeepSeek-AI}},
  title = {{DeepSeek-R1}: Incentivizing Reasoning Capability in {LLMs} via Reinforcement Learning},
  journal = {arXiv preprint arXiv:2501.12948},
  year = {2025}
}

@misc{deepseek2026updates,
  author = {{DeepSeek}},
  title = {{API} update notes},
  year = {2026},
  url = {https://api-docs.deepseek.com/updates/}
}

@article{yang2025qwen3,
  author = {Yang, An and Li, Anfeng and Yang, Baosong and others},
  title = {{Qwen3} Technical Report},
  journal = {arXiv preprint arXiv:2505.09388},
  year = {2025}
}

@article{qwen2025vl,
  author = {{Qwen Team}},
  title = {{Qwen3-VL} Technical Report},
  journal = {arXiv preprint arXiv:2511.21631},
  year = {2025}
}

@inproceedings{stephan2025calculation,
  author = {Stephan, Andreas and Zhu, Dawei and A{\ss}enmacher, Matthias and Shen, Xiaoyu and Roth, Benjamin},
  title = {From Calculation to Adjudication: Examining {LLM} Judges on Mathematical Reasoning Tasks},
  booktitle = {Proceedings of the GEM Workshop},
  year = {2025}
}

@misc{redhat2026cards,
  author = {{Red Hat AI}},
  title = {Model cards with baseline evaluations of the unquantized models: {Qwen3-32B}, {Qwen3-8B}, {Qwen3-4B}, gemma-3-4b-it, and {Meta-Llama-3.1-405B-Instruct}},
  year = {2026},
  howpublished = {Hugging Face},
  url = {https://huggingface.co/RedHatAI}
}

@misc{meta2026llama,
  author = {{Meta}},
  title = {Llama 3.1 model card},
  year = {2026},
  url = {https://huggingface.co/meta-llama/Llama-3.1-8B-Instruct}
}

@misc{microsoft2026phi,
  author = {{Microsoft}},
  title = {Phi-4-mini-instruct model card},
  year = {2026},
  url = {https://huggingface.co/microsoft/Phi-4-mini-instruct}
}

@misc{google2026gemma,
  author = {{Google}},
  title = {Gemma 3 model card},
  year = {2026},
  url = {https://ai.google.dev/gemma/docs/core/model_card_3}
}

@article{ling2025tech,
  author = {{Ling Team}},
  title = {{LING} technical report},
  journal = {arXiv preprint arXiv:2503.05139},
  year = {2025}
}

@article{alkiek2025big,
  author = {Alkiek, Kenan and Jurgens, David and Vydiswaran, Vinod},
  title = {Big Reasoning with Small Models: Instruction Retrieval at Inference Time},
  journal = {arXiv preprint arXiv:2510.13935},
  year = {2025}
}

@misc{openai2026gpt56,
  author = {{OpenAI}},
  title = {{GPT-5.6} release report},
  year = {2026},
  url = {https://openai.com/index/gpt-5-6/}
}

@misc{openai2026astra,
  author = {{OpenAI}},
  title = {{GPT-6 Astra} release report},
  year = {2026},
  url = {https://openai.com/index/gpt-6-astra/}
}

@article{gilardi2023chatgpt,
  author = {Gilardi, Fabrizio and Alizadeh, Meysam and Kubli, Ma{\"e}l},
  title = {{ChatGPT} Outperforms Crowd Workers for Text-Annotation Tasks},
  journal = {Proceedings of the National Academy of Sciences},
  volume = {120},
  number = {30},
  pages = {e2305016120},
  year = {2023}
}

@article{bucher2024finetuned,
  author = {Bucher, Martin Juan Jos{\'e} and Martini, Marco},
  title = {Fine-Tuned `Small' {LLMs} (Still) Significantly Outperform Zero-Shot Generative {AI} Models in Text Classification},
  journal = {arXiv preprint arXiv:2406.08660},
  year = {2024}
}

@article{chen2023frugalgpt,
  author = {Chen, Lingjiao and Zaharia, Matei and Zou, James},
  title = {{FrugalGPT}: How to Use Large Language Models While Reducing Cost and Improving Performance},
  journal = {arXiv preprint arXiv:2305.05176},
  year = {2023}
}

@inproceedings{ding2024hybrid,
  author = {Ding, Dujian and Mallick, Ankur and Wang, Chi and Sim, Robert and Mukherjee, Subhabrata and R{\"u}hle, Victor and Lakshmanan, Laks V. S. and Awadallah, Ahmed Hassan},
  title = {Hybrid {LLM}: Cost-Efficient and Quality-Aware Query Routing},
  booktitle = {International Conference on Learning Representations},
  year = {2024}
}

@article{ong2024routellm,
  author = {Ong, Isaac and Almahairi, Amjad and Wu, Vincent and Chiang, Wei-Lin and Wu, Tianhao and Gonzalez, Joseph E. and Kadous, M. Waleed and Stoica, Ion},
  title = {{RouteLLM}: Learning to Route {LLMs} with Preference Data},
  journal = {arXiv preprint arXiv:2406.18665},
  year = {2024}
}

\appendix

\section{Evaluation Details}
\label{app:details}

\paragraph{Benchmarks.} The evaluation sets contain 14,042 items for MMLU, 817 for TruthfulQA, 198 for GPQA Diamond, 1,172 for ARC-Challenge, and 254 for AQuA-RAT.

\paragraph{Aggregation.} The frontier median on each benchmark is taken over the frontier scores evaluated in this work: three on GPQA Diamond and six on each of the other 12 benchmarks. All derived numbers are computed from the scores shown in Table~\ref{tab:main}, and medians are rounded to one decimal place before differences are taken.

\paragraph{TruthfulQA.} Our TruthfulQA scores are the accuracy of the selected option. The published scores from sources h and j (Table~\ref{tab:sources}) use the MC2 metric, which measures the probability assigned to true answers. They are marked with $^{\ast}$ and excluded from the counts and mean differences in Tables~\ref{tab:rep} and~\ref{tab:small}.

\paragraph{Uncertainty.} The 95\% Wilson intervals for Jev are [67.7, 79.8] on GPQA Diamond, [78.0, 87.2] on AQuA-RAT, and [96.6, 98.3] on ARC-Challenge. The interval for Claude Opus 5.5 on GPQA Diamond is [69.3, 81.2].

\section{Additional Results}
\label{app:additional}

\paragraph{Gap to the frontier.} Table~\ref{tab:gap} gives the numbers behind Figure~\ref{fig:gap}.

\begin{table}[h]
  \centering
  \footnotesize
  \setlength{\tabcolsep}{2.5pt}
  \begin{tabular}{lccccc}
    \toprule
    \textbf{Benchmark} & \textbf{Jev} & \textbf{Median} & \textbf{Range} & \textbf{\boldmath $\Delta$} & \textbf{Rank} \\
    \midrule
    \textbf{TruthfulQA} & 87.5 & 83.0 & 76.0--97.9 & +4.5 & 3/7 \\
    \textbf{ARC-Challenge} & 97.6 & 95.5 & 91.4--97.5 & +2.1 & 1/7 \\
    \textbf{MMLU-Redux} & 95.0 & 93.2 & 90.0--94.3 & +1.8 & 1/7 \\
    \textbf{WinoGrande} & 91.8 & 92.1 & 70.2--97.1 & $-$0.3 & 4/7 \\
    \textbf{MMLU} & 91.3 & 92.5 & 87.5--94.4 & $-$1.2 & 5/7 \\
    \textbf{MMLU-ProX} & 83.2 & 84.5 & 68.9--87.6 & $-$1.3 & 5/7 \\
    \textbf{MMMLU} & 85.0 & 88.6 & 85.0--92.6 & $-$3.6 & 6/7 \\
    \textbf{HellaSwag} & 89.3 & 93.8 & 88.0--97.5 & $-$4.5 & 6/7 \\
    \textbf{MMLU-Pro} & 82.3 & 88.1 & 84.5--93.8 & $-$5.8 & 7/7 \\
    \textbf{AQuA-RAT} & 83.1 & 91.0 & 89.0--94.1 & $-$7.9 & 7/7 \\
    \textbf{C-Eval} & 85.3 & 93.6 & 83.1--94.7 & $-$8.3 & 6/7 \\
    \textbf{GPQA Diamond} & 74.2 & 83.1 & 75.8--92.4 & $-$8.9 & 4/4 \\
    \textbf{MathQA} & 75.0 & 92.7 & 85.8--94.4 & $-$17.7 & 7/7 \\
    \bottomrule
  \end{tabular}
  \caption{Jev against the six frontier LLMs: the median and range of their scores, the difference $\Delta$ between Jev and the median, and the rank of Jev among the evaluated models. On GPQA Diamond, only the three frontier LLMs evaluated in this work are counted.}
  \label{tab:gap}
\end{table}

\section{Sources of Published Scores}
\label{app:sources}

Tables~\ref{tab:rep} and~\ref{tab:small} give the source of every published score of the representative and small LLMs, and Table~\ref{tab:sources} lists the sources and their protocols. Scores without a source are evaluated in this work. All sources were accessed on September 30, 2026. The published DeepSeek-V3 scores on MMLU-Pro and MMLU-Redux are those of the V3-0324 checkpoint, and its other published scores are those of the original V3.

\begin{table*}[t]
  \centering
  \small
  \setlength{\tabcolsep}{4pt}
  \resizebox{\textwidth}{!}{%
  \begin{tabular}{lcccccccc}
    \toprule
    \textbf{Benchmark} & \textbf{Jev} & \textbf{GPT-4o} & \textbf{Claude 3.5 Sonnet} & \textbf{DeepSeek-R1} & \textbf{DeepSeek-V3} & \textbf{Qwen3-32B} & \textbf{Qwen2.5-72B} & \textbf{Llama-405B} \\
    \midrule
    \textbf{MMLU-Pro} & 82.3 & \textit{72.6}$^{a}$ & \textit{78.0}$^{a}$ & \textbf{\textit{84.0}}$^{c}$ & \textit{81.2}$^{b}$ & \textit{79.1}$^{e}$ & \textit{71.6}$^{a}$ & \textit{73.3}$^{a}$ \\
    \textbf{MMLU-Redux} & \textbf{95.0} & \textit{88.0}$^{a}$ & \textit{88.9}$^{a}$ & \textit{92.9}$^{c}$ & \textit{90.4}$^{e}$ & \textit{90.9}$^{d}$ & \textit{85.6}$^{a}$ & \textit{86.2}$^{a}$ \\
    \textbf{MMLU} & \textbf{91.3} & \textit{87.2}$^{a}$ & \textit{88.3}$^{a}$ & \textit{90.8}$^{c}$ & \textit{88.5}$^{a}$ & \textit{81.0}$^{h}$ & \textit{85.3}$^{a}$ & \textit{88.6}$^{a}$ \\
    \textbf{GPQA Diamond} & \textbf{74.2} & \textit{49.9}$^{a}$ & \textit{65.0}$^{a}$ & \textit{71.5}$^{c}$ & \textit{59.1}$^{a}$ & \textit{68.4}$^{d}$ & \textit{49.0}$^{a}$ & \textit{51.1}$^{a}$ \\
    \textbf{C-Eval} & 85.3 & \textit{76.0}$^{a}$ & \textit{76.7}$^{a}$ & \textbf{\textit{91.8}}$^{c}$ & \textit{86.5}$^{a}$ & \textit{87.3}$^{d}$ & \textit{86.1}$^{a}$ & \textit{61.5}$^{a}$ \\
    \textbf{HellaSwag} & 89.3 & \textbf{92.6} & 88.6 & 91.5 & 89.2 & \textit{71.1}$^{h}$ & 87.2 & \textit{88.3}$^{h}$ \\
    \textbf{WinoGrande} & \textbf{91.8} & 87.1 & 85.0 & 89.7 & 84.9 & \textit{69.8}$^{h}$ & 82.7 & \textit{87.2}$^{h}$ \\
    \textbf{TruthfulQA} & \textbf{87.5} & 86.0 & 81.5 & 84.5 & 82.1 & \textit{58.6}$^{h\ast}$ & 80.1 & 82.4 \\
    \textbf{ARC-Challenge} & \textbf{97.6} & 96.1 & 93.7 & 96.6 & 95.3 & \textit{69.0}$^{h}$ & 92.7 & \textit{96.9}$^{i}$ \\
    \textbf{MathQA} & 75.0 & 86.0 & 90.0 & 88.8 & 87.8 & \textbf{91.9} & 85.7 & 84.7 \\
    \textbf{AQuA-RAT} & 83.1 & 88.0 & 86.0 & \textbf{88.8} & 86.3 & 88.6 & \textit{83.7}$^{g}$ & 81.0 \\
    \textbf{MMLU-ProX} & \textbf{83.2} & \textit{71.1}$^{f}$ & 68.7 & \textit{74.6}$^{f}$ & \textit{70.8}$^{f}$ & \textit{69.2}$^{f}$ & \textit{58.7}$^{f}$ & 59.8 \\
    \textbf{MMMLU} & 85.0 & 77.9 & 81.1 & \textbf{86.8} & 79.4 & \textit{80.6}$^{d}$ & \textit{76.9}$^{d}$ & 75.0 \\
    \midrule
    Jev ahead / compared & & 10/13 & 11/13 & 7/13 & 10/13 & 9/12 & 10/13 & 12/13 \\
    Mean difference & & +4.8 & +3.8 & $-$0.9 & +3.0 & +7.2 & +7.3 & +8.0 \\
    \bottomrule
  \end{tabular}}
  \caption{Jev and the representative LLMs. \textbf{Bold} marks the best score on each benchmark. Scores in \textit{italics} are published, and their superscripts refer to the sources in Table~\ref{tab:sources}; the other scores are evaluated in this work. $^{\ast}$MC2 metric, excluded from the last two rows. Qwen2.5-72B and Llama-405B are the Instruct models.}
  \label{tab:rep}
\end{table*}

\begin{table*}[t]
  \centering
  \small
  \setlength{\tabcolsep}{5pt}
  \begin{tabular}{lccccccc}
    \toprule
    \textbf{Benchmark} & \textbf{Jev} & \textbf{Qwen3-8B} & \textbf{Qwen3-4B} & \textbf{Llama-3.1-8B} & \textbf{Phi-4-mini} & \textbf{Gemma-3-4B} & \textbf{Mistral-7B} \\
    \midrule
    \textbf{MMLU-Pro} & \textbf{82.3} & \textit{74.6}$^{e}$ & \textit{70.4}$^{e}$ & \textit{48.3}$^{i}$ & \textit{52.8}$^{j}$ & \textit{43.6}$^{k}$ & \textit{31.5}$^{l}$ \\
    \textbf{MMLU-Redux} & \textbf{95.0} & \textit{87.5}$^{d}$ & \textit{83.7}$^{d}$ & \textit{61.7}$^{d}$ & 73.1 & 71.1 & \textit{58.4}$^{l}$ \\
    \textbf{MMLU} & \textbf{91.3} & \textit{72.0}$^{h}$ & \textit{66.8}$^{h}$ & \textit{69.4}$^{i}$ & \textit{67.3}$^{j}$ & \textit{58.4}$^{h}$ & \textit{63.5}$^{l}$ \\
    \textbf{GPQA Diamond} & \textbf{74.2} & \textit{62.0}$^{d}$ & \textit{55.9}$^{d}$ & \textit{32.8}$^{d}$ & 32.0 & \textit{30.8}$^{k}$ & 30.2 \\
    \textbf{C-Eval} & \textbf{85.3} & \textit{83.4}$^{d}$ & \textit{77.5}$^{d}$ & \textit{52.0}$^{d}$ & 61.4 & 59.5 & \textit{45.9}$^{l}$ \\
    \textbf{HellaSwag} & \textbf{89.3} & \textit{56.5}$^{h}$ & \textit{52.8}$^{h}$ & 73.5 & \textit{69.1}$^{j}$ & \textit{75.0}$^{h}$ & \textit{75.8}$^{l}$ \\
    \textbf{WinoGrande} & \textbf{91.8} & \textit{66.0}$^{h}$ & \textit{58.4}$^{h}$ & 64.7 & \textit{67.0}$^{j}$ & \textit{70.3}$^{h}$ & \textit{77.6}$^{l}$ \\
    \textbf{TruthfulQA} & \textbf{87.5} & \textit{53.2}$^{h\ast}$ & \textit{51.8}$^{h\ast}$ & 69.2 & \textit{66.4}$^{j\ast}$ & \textit{51.9}$^{h\ast}$ & 64.7 \\
    \textbf{ARC-Challenge} & \textbf{97.6} & \textit{61.7}$^{h}$ & \textit{50.2}$^{h}$ & \textit{83.4}$^{i}$ & \textit{83.7}$^{j}$ & \textit{56.6}$^{h}$ & \textit{72.2}$^{l}$ \\
    \textbf{MathQA} & 75.0 & \textbf{92.3} & \textit{90.0}$^{m}$ & 81.2 & 82.8 & 84.3 & 76.7 \\
    \textbf{AQuA-RAT} & 83.1 & \textbf{86.3} & 83.0 & \textit{61.5}$^{g}$ & 72.8 & 72.2 & 61.0 \\
    \textbf{MMLU-ProX} & \textbf{83.2} & \textit{60.0}$^{f}$ & \textit{56.6}$^{f}$ & \textit{28.6}$^{f}$ & \textit{28.4}$^{f}$ & \textit{25.1}$^{f}$ & 31.1 \\
    \textbf{MMMLU} & \textbf{85.0} & \textit{74.4}$^{d}$ & \textit{69.8}$^{d}$ & \textit{44.4}$^{d}$ & 59.8 & 57.1 & 41.9 \\
    \midrule
    Jev ahead / compared & & 10/12 & 11/12 & 12/13 & 11/12 & 11/12 & 12/13 \\
    Mean difference & & +13.0 & +18.2 & +26.9 & +23.6 & +27.4 & +30.0 \\
    \bottomrule
  \end{tabular}
  \caption{Jev and the small LLMs. \textbf{Bold} marks the best score on each benchmark. Scores in \textit{italics} are published, and their superscripts refer to the sources in Table~\ref{tab:sources}; the other scores are evaluated in this work. $^{\ast}$MC2 metric, excluded from the last two rows. Llama-3.1-8B and Phi-4-mini are the Instruct models, Gemma-3-4B is the IT model, and Mistral-7B is the v0.3 base model.}
  \label{tab:small}
\end{table*}

\begin{table*}[t]
  \centering
  \small
  \begin{tabular}{llp{0.53\textwidth}}
    \toprule
    \textbf{Key} & \textbf{Source} & \textbf{Models and protocol} \\
    \midrule
    a & DeepSeek-V3 report \citep{deepseek2024v3} & Chat models: GPT-4o (0513), Claude 3.5 Sonnet (1022), Qwen2.5-72B-Instruct, Llama-3.1-405B-Instruct, and the original DeepSeek-V3. \\
    b & DeepSeek update notes \citep{deepseek2026updates} & DeepSeek-V3-0324 on MMLU-Pro; DeepSeek-V4.1-Flash on GPQA Diamond. \\
    c & DeepSeek-R1 report \citep{deepseek2025r1} & Original DeepSeek-R1. \\
    d & Qwen3 report \citep{yang2025qwen3} & Qwen3 in thinking mode; Llama-3.1-8B-Instruct as a baseline; MMMLU over 14 languages. \\
    e & Qwen3-VL report \citep{qwen2025vl} & Text-only Qwen3 baselines in thinking mode; DeepSeek-V3-0324 on MMLU-Redux. \\
    f & MMLU-ProX \citep{xuan2025mmluprox} & 29 languages, full set, 5-shot with chain-of-thought. \\
    g & \citet{stephan2025calculation} & AQuA-RAT test set, 4-shot with chain-of-thought. \\
    h & Red Hat AI model cards \citep{redhat2026cards} & Unquantized baselines scored by log-likelihood: MMLU 5-shot, ARC-Challenge 25-shot, HellaSwag 10-shot, WinoGrande 5-shot, TruthfulQA MC2 zero-shot. \\
    i & Llama 3.1 model card \citep{meta2026llama} & MMLU 5-shot, MMLU-Pro 5-shot with chain-of-thought, ARC-Challenge zero-shot. \\
    j & Phi-4-mini model card \citep{microsoft2026phi} & MMLU 5-shot, MMLU-Pro zero-shot with chain-of-thought, ARC-Challenge 10-shot, HellaSwag and WinoGrande 5-shot, TruthfulQA MC2 10-shot. \\
    k & Gemma 3 model card \citep{google2026gemma} & Instruction-tuned model, zero-shot. \\
    l & LING report \citep{ling2025tech} & Mistral-7B-v0.3 base model, 5-shot. \\
    m & \citet{alkiek2025big} & Qwen3-4B-Instruct on MathQA, zero-shot. \\
    n & GPT-5.6 report \citep{openai2026gpt56} & GPT-5.6 Sol on GPQA Diamond. \\
    o & GPT-6 Astra report \citep{openai2026astra} & Gemini 3.8 Flash on GPQA Diamond. \\
    \bottomrule
  \end{tabular}
  \caption{Sources and protocols of the published scores. Keys n, o, and b also cover the vendor-reported GPQA Diamond scores of GPT-5.6 Sol, Gemini 3.8 Flash, and DeepSeek-V4.1-Flash in Table~\ref{tab:main}.}
  \label{tab:sources}
\end{table*}

\end{document}